\documentclass[letterpaper, 10 pt, conference]{ieeeconf}  
\IEEEoverridecommandlockouts                              
\usepackage{graphicx} 
\usepackage{amsmath} 
\usepackage{amssymb}  
\usepackage{multirow}
\usepackage{color}
\usepackage{subfig}
\newcommand{\etal}{\textit{et al.~}}
\usepackage[table]{xcolor}

\title{\LARGE \bf Emotion Recognition Using Fusion of Audio and Video Features}
\author{Juan D. S. Ortega, Patrick Cardinal and Alessandro L. Koerich
\thanks{*This work was not supported by any organization}
\thanks{Juan David Silva Ortega, Patrick Cardinal and Alessandro Lameiras Koerich are with Department of Software and IT Engineering, \'{E}cole de Technologie Sup\'{e}rieure, University of Qu\'{e}bec, H3C 1K9, Montr\'{e}al, QC, Canada.
        {\tt\footnotesize juan-david.silva-ortega.1@ens.etsmtl.ca, patrick.cardinal@etsmtl.ca, alessandro.koerich@etsm tl.ca}}%
}

\begin{document}
\maketitle
\thispagestyle{empty}
\pagestyle{empty}

\begin{abstract}
In this paper we propose a fusion approach to continuous emotion recognition that combines visual and auditory modalities in their representation spaces to predict the arousal and valence levels. The proposed approach employs a pre-trained convolution neural network and transfer learning to extract features from video frames that capture the emotional content. For the auditory content, a minimalistic set of parameters such as prosodic, excitation, vocal tract, and spectral descriptors are used as features. The fusion of these two modalities is carried out at a feature level, before training a single support vector regressor (SVR) or at a prediction level, after training one SVR for each modality. The proposed approach also includes preprocessing and post-processing techniques which contribute favorably to improving the concordance correlation coefficient (CCC). Experimental results for predicting spontaneous and natural emotions on the RECOLA dataset have shown that the proposed approach takes advantage of the complementary information of visual and auditory modalities and provides CCCs of 0.749 and 0.565 for arousal and valence, respectively.  \end{abstract}

\section{Introduction}
Understand the emotional state of people is a phenomenon that has attracted the attention and also fascinate researchers from different branches of science for a while. Psychologists, psychiatrists, neuroscientists, and computer scientists constantly try to untangle the combination of variables that best describes an emotional state. The circumplex model, proposed by Russell \cite{Russell1989} generates a circular representation of the emotional space. This model suggests that there are two independent neurophysiological systems and their combination generates a representation of an emotional state. This model is represented by coordinated axis of two dimensions, where the vertical axis represents the alertness/activeness level (arousal) in the emotion and the horizontal axis represents the pleasure/displeasure level (valence). Both dimensions can be normalized in the range between -1 and +1. A vast mixture of emotions can be described with the linear combination of arousal and valence.

In the last years, several works attempt to predict valence and arousal using machine learning algorithms \cite{Eyben2012, Ringeval2015, Weber2016, Cardinal2015}. Eyben \etal \cite{Eyben2012} proposed a fully automatic audiovisual recognition approach based on Long Short-Term Memory (LSTM) modeling of word-level audio and video features. Evaluations carried out on the SEMAINE dataset have shown how acoustic, linguistic, and visual features contribute to the recognition of different affective dimensions. Ringeval \etal \cite{Ringeval2015} investigated the relevance of using machine learning algorithms to integrate contextual information in the modeling to automatically predict emotion from several raters in continuous time domains. Evaluations carried out on the RECOLA dataset have achieved concordance correlation coefficient (CCC) of 0.804 and 0.528 for arousal and valence reespectively. Weber \etal \cite{Weber2016} proposed to improve the performance of the multimodal prediction with low-level features by adding high-level geometry-based features as well as to fuse the unimodal predictions trained on each training subject before performing the multimodal fusion. The results show high-level features improve the performance of the multimodal prediction of arousal and that the subject's fusion works well in unimodal prediction but generalizes poorly in multimodal prediction, particularly on valence. Ding \etal \cite{Ding2016} presented the fusion of facial expression recognition and audio emotion recognition at score level. For facial emotion recognition, a pre-trained deep Convolutional Neural Network (CNN) fine-tuned on FER2013 dataset for feature extraction and kernel SVM, logistic regression and partial least squares are used as classifiers. For audio emotion recognition, a deep LSTM Recurrent Neural Network (LSTM-RNN) is trained directly on the Emotion Recognition in the Wild 2016 challenge dataset. Experimental results have shown that the proposed approach achieved state-of-the-art performance with an overall accuracy of 53.9{\%} on the test dataset . Yan \etal \cite{Yan2018} presented a transductive deep transfer learning architecture based on VGGface16 CNN which is used to jointly learn optimal nonlinear discriminative features from non-frontal facial expressions. Cross-dataset experiments on BU-3DEF and Multi-PIE datasets have shown that the proposed approach outperforms the state-of-the-art cross-database facial expression recognition methods. Yan \etal \cite{Yan2018a} propose a multi-cue fusion emotion recognition framework by modeling human emotions from three complementary cues: facial texture, facial landmark action and audio signal, and then fusing them together. Models are fused at both feature level and decision level and the experimental results on AFEW and CHEAVD datasets have shown the effectiveness of the proposed approach. Tzirakis \etal \cite{Tzirakis2017} proposed an emotion recognition system using auditory and visual modalities. They utilized a pre-trained CNN to extract features from the speech and a deep residual network to extract features from the visual modality.

\begin{figure*}[hbtp!]
  \centering
  \begin{tabular}{ccc}
\includegraphics[width=0.25\textwidth]{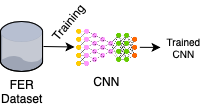} &\hspace{4pt} &  
\includegraphics[width=0.65\textwidth]{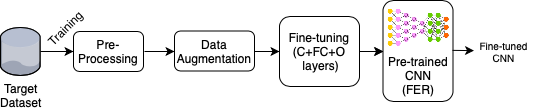} \\
(a) &  & (b) \\
\end{tabular}
  \caption{(a) Pre-training a CNN in a source dataset; (b) Fine-tuning the CNN on the target dataset.}
\label{approach}  
\end{figure*}

\begin{figure}[hbtp!]
  \centering
\includegraphics[width=0.48\textwidth]{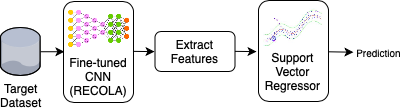} \\
  \caption{Using the fine-tuned CNN to extract features and train an SVR.}
\label{approach2}  
\end{figure}

In this paper we propose an approach based on transfer learning and multimodal fusion for the problem of continuous emotion recognition in video. Facial features are extracted from video frames by a pre-trained CNN and such features are fused with handcrafted audio features extracted from the subjects' voice. We evaluate two fusion schemes, early and late as well as the several pre- and post-processing techniques to improve the prediction of arousal and valence levels. The main contributions of this paper are: (i) a compact representation learned from raw data that is comparable to the state-of-the-art; (ii) a multimodal fusion approach of a learned representation and handcrafted features that yields to a compact representation and that achieved CCCs that are comparable to the state-of-the-art as well.

This paper is organized as follows. Section \ref{sec:prop} presents the proposed approach for dealing with raw video, the proposed fusion schemes and the pre- and post-processing techniques. Section \ref{sec:exp} presents the experimental results for predicting valence and arousal on RECOLA dataset. Finally, in the last section we present our conclusions and perspectives of future work.
\section{Proposed Approach}
\label{sec:prop} 
Fig.~\ref{approach} shows a general structure of the proposed approach to predict arousal and valence levels. We start with transfer learning, a technique that allows us to import information from another model that was trained on a similar task. The idea is to use a source dataset (FER) which provides a large number of face images with emotional content to train a CNN. Transfer learning allows us to use such a pre-trained CNN in our target task instead of training a CNN initialized with random parameters. The target dataset (RECOLA) is then used to fine-tune some parameters (layers) of such a CNN. The target dataset provides video data, audio, and physiological signals as well as handcrafted features extracted from these modalities. Video frames that allow us to build a regressor to predict levels of arousal and valence of a subject. Because of the nature of the datasets and the strong relation between the tasks (emotional prediction or classification), we can say empirically that even if we train two network models, one for each dataset, the feature space should be correlated.

However, before fine-tuning the CNN on the target dataset, we propose to use certain preprocessing techniques to improve the quality of the training data, such as: (i) frame suppression to discard video frames where face is not detected; (ii) frame quality selection to filter detected face are far from being frontal face images; (iii) delay compensation to realign labels and frames to compensate the reaction lag of annotators \cite{6681412}. Finally, as the amount of training data is critical to train a CNN, given its high number of trainable parameters, a data augmentation strategy is used to address this issue.

\begin{figure*}[hbtp!]
\centering
\includegraphics[width=0.90\textwidth]{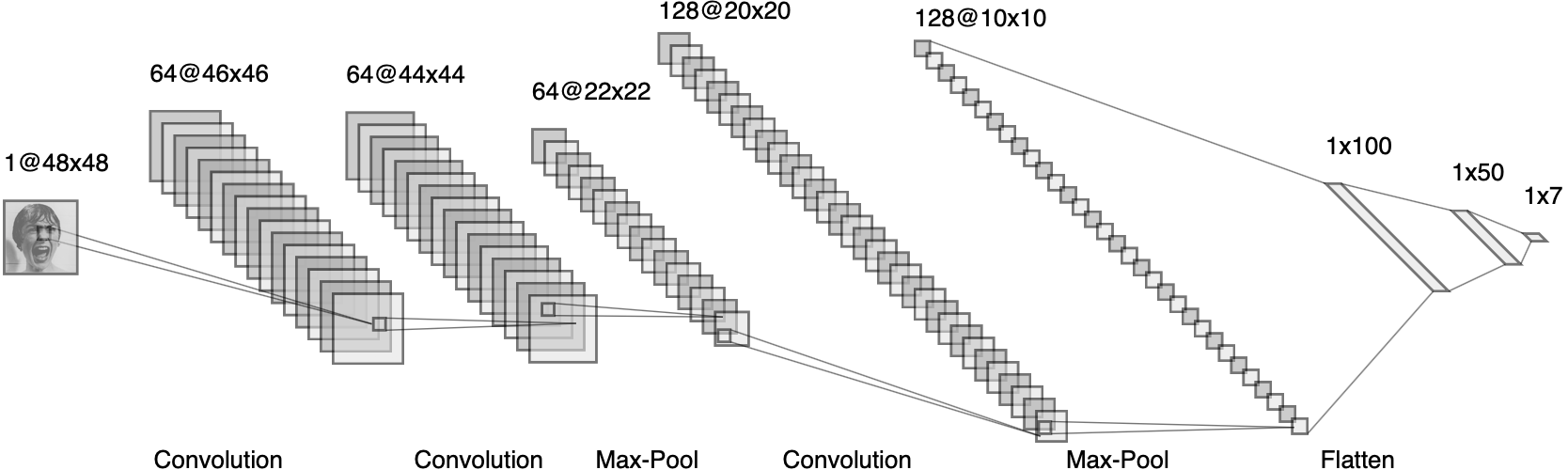}
\caption{The CNN architecture for the FER dataset with three convolutional layers (CL) and two full connected (FC) layers.}
\label{fig:CNN}  
\end{figure*}

The proposed CNN shown in Fig.~\ref{fig:CNN}, is based on the architecture presented by Sun \etal \cite{Sun2016}, which achieved 67.8\% of accuracy on the test set of FER dataset. We have introduced some modifications on such an architecture in an attempt to improving the learned representation. We have changed the number of convolutional layers and fully-connect layers, to reduce the number of parameters of the network and have a best trade-off between the complexity and the amount of data available for training. The proposed architectures are shown in Table~\ref{architecturea}.

\begin{table}
\caption{Three CNN architectures based on \cite{Sun2016}: (A) With less CL and FC layers; (B) With a reduced number of CL and FC layers; (C) With more CLs.}
\renewcommand{\arraystretch}{1.2}
\label{architecturea}
\centering
\begin{tabular}{|c|c|c|c|}
\hline
\textbf{Layers}                 & \multicolumn{3}{c|}{\textbf{Architecture}}    \\ \cline{2-4}
& A & B & C\\
\hline
Zero-Padding(2D)     &   1$\times$1 &  & 1$\times$1 \\ \cline{2-2}\cline{4-4}\cline{3-3}
Convolution     &64$\times$3$\times$3 & 64$\times$3$\times$3  & 256$\times$3$\times$3 \\ \cline{4-4}\cline{2-2}\cline{3-3}
Zero-Padding(2D)     &    &  & 1$\times$1 \\ \cline{2-2}\cline{4-4}
Max Pooling            & 2$\times$2 &  &      \\ \cline{2-2}
Zero-Padding(2D)     &   1$\times$1 &  &  \\ \cline{4-4} \cline{3-3} \cline{2-2}
Convolution     & 128$\times$3$\times$3 & 64$\times$3$\times$3  & 256$\times$3$\times$3 \\ \cline{2-2}\cline{3-3}\cline{4-4}
Max Pooling            & 2$\times$2  & 2$\times$2 &     \\ \cline{4-4} \cline{2-2} \cline{3-3}
Zero-Padding(2D)     &   1$\times$1 &  & 1$\times$1 \\ \cline{4-4} \cline{2-2}\cline{3-3}
Convolution     & 256$\times$3$\times$3 & 128$\times$3$\times$3 & 256$\times$3$\times$3 \\ \cline{2-2} \cline{4-4}\cline{3-3}
Max Pooling            & 2$\times$2   & 2$\times$2 & 2$\times$2   \\ \cline{4-4} \cline{2-2}\cline{3-3}
Zero-Padding(2D)     &  1$\times$1 &  & 1$\times$1  \\ \cline{4-4} \cline{2-2}
Convolution     & 256$\times$3$\times$3 &  & 256$\times$3$\times$3 \\ \cline{4-4} \cline{2-2}
Zero-Padding(2D)     &    &  & 1$\times$1 \\ \cline{4-4}
Convolution     &  &    & 256$\times$3$\times$3 \\ \cline{4-4}
Zero-Padding(2D)     &     &  & 1$\times$1 \\ \cline{4-4}
Convolution     &  &    & 256$\times$3$\times$3 \\ \cline{4-4} \cline{2-2}
Max Pooling            & 2$\times$2 &  & 2$\times$2    \\ \cline{4-4}\cline{2-2}\cline{3-3}
Fully Connected & 1024       & 100 & 1024    \\ \cline{4-4}\cline{3-3} \cline{2-2}
Fully Connected &    & 50 & 1024    \\ \cline{4-4} \cline{2-2} \cline{3-3}
Fully Connected & 7     & 7 & 7  \\\hline
\hline
\textbf{\# of Parameters} & 3.3M & 1.4M & 41.7M\\ \hline     
\end{tabular}
\end{table}

Before fine-tuning these architectures for our target task, which is the regression problem of predicting continuous emotions, we have replaced the output layer by a single neuron to obtain a real value at the output. For the fine-tuning process, we started by freezing (making them non-trainable) all convolutional layers and fine-tuning only the fully connected layers. Then, we have progressively unfrozen the convolutional layers to evaluate if the network could improve the representation learned on the FER dataset to the target dataset (RECOLA). Besides fine-tuning the CNNs, we have also trained CNNs end-to-end, which means that they were initialized randomly. Finally, as illustrated in Fig.~\ref{approach2}, we have used such CNNs to generate feature vectors from video frames by taking the representation provided by the fully connected layers and use them as the input of another learning algorithm, or to fuse such feature vectors with feature vectors of another modality as a complementary representation of the emotional state.

For the multimodal fusion, we propose to use the extended Geneva Minimalistic Acoustic Parameter Set (eGeMAPS), which are low level descriptors that cover spectral, cepstral, prosodic and voice quality information of the voice record. Such features have been used in the RECOLA baseline \cite{Valstar2016} together with other modalities. Table~\ref{contribution} shows the average contribution to the final prediction of video appearance (V$_{\text {App}}$) and geometric (V$_{\text {Geo}}$), electrocardiogram (ECG), heart rate and heart rate variability (HR), electrodermal activity (EDA), skin conductance level (SCL) and resistance (SCR). The average audio contribution is 0.320 which is the highest among all modalities. Therefore, we choose such a modality to build our multimodal approach.

\begin{table}[htpb!]
\centering
\caption{RECOLA baseline: average valence-arousal contribution of each modality in the final prediction.}
\renewcommand{\arraystretch}{1.2}
\label{contribution}
        \begin{tabular}{|c|c|c|c|c|c|c|c|}
 		\hline
 			\multirow{2}{*}{\bf Audio} & \multicolumn{2}{c|}{\textbf{Video}} & \multirow{2}{*}{\bf ECG} & \multirow{2}{*}{\bf HR} & \multirow{2}{*}{\bf EDA} & \multirow{2}{*}{\bf SCL} & \multirow{2}{*}{\bf SCR} \\
 			\cline{2-3}
 			 & {\bf App} & {\bf Geo} &  &  &  &  &  \\
 	  \hline
 			0.320 & 0.085 & 0.170 & 0.03 & 0.115 & 0.035 & 0.075 & 0.170 \\
 	  \hline
 		\end{tabular}
 \end{table}
 
Considering that we use CNNs as feature extractors, we must define how to use such features to make predictions. Valstar \etal \cite{Valstar2016} show how the features of the RECOLA baseline perform well with an SVR algorithm. Therefore, we also use an SVR and compare its performance with that of the CNN. Also, tuning an SVR in the fusion approach is less complex than tuning a CNN that takes into account multiple modalities, because an SRV requires adjusting only two hyperparameters instead of a large number of weights. The grid search strategy can find a suitable combination of $C$ and $\epsilon$ given a range of values for both hyperparameters. An SVR algorithm can either predicts the arousal and valence levels by using single or multiple data modalities. Therefore, different fusion techniques can be used for such an aim.

\subsection{Fusion}
\begin{figure*}[hbtp!]
  \centering
  \begin{tabular}{ccc}
\includegraphics[width=0.32\textwidth]{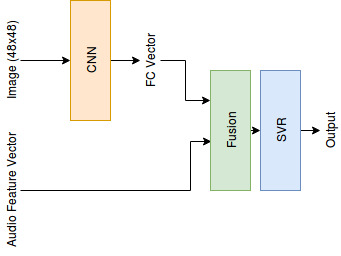} &\hspace{5pt} &  
\includegraphics[width=0.32\textwidth]{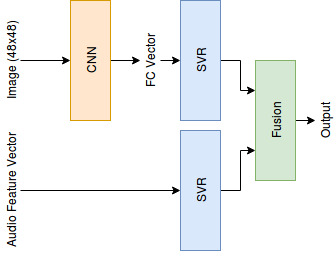}\\
(a) & & (b)
\end{tabular}
  \caption{(a) Early fusion: video and audio features are concatenated and used to train an SVR; (b) Late fusion: video and audio features are used to train two SVRs and the predictions of both are fused to generate an output.}
\label{cnn-svr}  
\end{figure*}

We have evaluated the early and the late schemes for fusing video and audio features. Early fusion, as shown in Fig.~\ref{cnn-svr}a, stacks all representations as a single multimodal vector and train a single SVR. Therefore, the assumption is that the model will learn from the richest source of information \cite{Snoek2005,Zhu2017,tannugi2019memory}. Late fusion, as illustrated in Fig.~\ref{cnn-svr}b uses unimodal vectors to train different SVRs and the predictions provided by each regressor (one by modality) and further fused by averaging them into one single prediction. Late fusion is more difficult to tune because we must train and find the hyper-parameters of at least two models. On the other hand, late fusion is more flexible than early fusion because each model has its own representation space and its own hyper-parameters.

\subsection{Post-Processing}
The propose approach does not takes into account the correlation that exist between consecutive frames. Therefore, the predictions of our model may have a noisy behavior. To mitigate such a problem, post-processing techniques such as median filtering, scaling factor, and centering \cite{Trigeorgis2016} allow the enhancement of the predictions of the model. Median filtering smooths our predictions by filtering the 1D output array with a window ranging between 0.4 and 8 seconds. The scaling factor $\beta$ can be calculated by Eq.~\ref{scaling1} as the ratio of the gold standard ($GS$) and the prediction ($Pr$) over the training set. The prediction on the development set (Eq.~\ref{scaling2}) is simply multiplied by this factor with the aim of rescaling the output.

\begin{equation}
\beta_{tr} =\frac{{GS} _{tr}}{{Pr}_{tr}}
\label{scaling1}
\end{equation}

\begin{equation}
{Pr}^\prime_{dev}=\beta_{tr} *{Pr}_{dev}
\label{scaling2}
\end{equation}

Finally, for centering the prediction distribution and align the ranges between it, we subtract the mean of the gold standard $\bar{y}_{GS}$ by the prediction $y$: 

\begin{equation}
y^\prime = y-\bar{y}_{GS}
\label{meanbias}
\end{equation}

\noindent where $y^\prime$ is the corrected prediction value.

\section{Experimental Results}
\label{sec:exp} 
The target dataset used to evaluate the proposed approach is RECOLA, which is a multimodal dataset composed of 9.5 hours of audio, visual, and physiological recordings captured from 46 French-speaking participants. However, only the data of 18 speakers is publicly available from which nine speakers compose the training set and another nine speakers compose the developing set. We only used the audio and the visual modalities, because we aim a non-invasive approach for emotion detection.

Two metrics are used to evaluate the behavior of the different CNN architectures on the target dataset. The Mean Absolute Error (MAE), which offers equity over all the absolute differences between golden standard and predictions and is defined as:

\begin{equation}
\text{MAE}=\frac{1}{n}\sum_{j=1}^{n}\left | y_{j}-y_{GS_j} \right |
\label{MAE}
\end{equation}

\noindent where $n$ represents the total number of samples, $y_{GS}$ is the gold standard and $y$ is the prediction.

The second metric is the Concordance Correlation Coefficient (CCC) which is defined in Eq.~\ref{CCC}. It combines the Pearson correlation coefficient ($\rho$) with the squared difference between the mean of the predictions ($\mu_y$) and the mean of the golden standard ($\mu_{{y}_{GS}}$). This metric measures the association between variables and penalizes the score even if the model predicts the emotion well, but it shifts the value.

\begin{equation}
\text{CCC} = \frac{2\rho\sigma_{y}\sigma_{{y}_{GS}}}{\sigma_{y}^{2}+\sigma_{{y}_{GS}}^{2}+(\mu_{y}-\mu_{{y}_{GS}})^{2}}
\label{CCC}
\end{equation}

\noindent where $\sigma^{2}_{y}$ and $\sigma^{2}_{{y}_{GS}}$ are the variance of the prediction and the gold standard, respectively.

\subsection{FER Dataset}
The FER dataset is used to pre-train the CNN architectures proposed in Section II. This dataset is composed of gray scale images of 48$\times$48 pixels that comprise seven acted emotions (disgust, anger, fear, joy, saddens, surprise, neutral). The FER dataset is split into 28,709 images for training, 3,589 for validation, and 3,589 for test. Table~\ref{parameters} shows the accuracy of the propose CNN architectures (Table~\ref{architecturea}) on the FER test set. The architecture B, which has less trainable parameters than architectures A and C, achieved the best accuracy of 67.7\%, which is comparable to the results reported in \cite{Ding2016,Goodfellow2013,Wan2016}. Therefore, for all further steps we adopted architecture B as our baseline.

\begin{table}[htpb!]
\caption{Accuracy for the three proposed architectures of Table~\ref{architecturea} on FER dataset.}
\renewcommand{\arraystretch}{1.2}
\label{parameters}
\begin{center}
\begin{tabular}{|c|c|c|} 
\hline
\multirow{2}{*}{\textbf{Architecture}} & \multicolumn{2}{c|}{\textbf{Accuracy (\%)}}  \\ 
\cline{2-3}
& \textbf{Train} & \textbf{Test} \\
\hline
A & 59.6 & 62.5 \\ 
\hline
B & \textbf{90.3} & \textbf{67.7} \\
\hline
C & 25.0 & 24.7 \\
\hline
\end{tabular}
\end{center}
\end{table}

\subsection{Transfer Learning}
Given the CNN pre-trained on FER dataset, we replace the output layer by a single neuron with linear activation to implement a regressor. Next, we freeze all convolutional layers (CL) and fine-tune (training) just the fully connected (FC) layers using the target dataset (RECOLA). Progressively, we unfreeze one convolutional layer per training session from the deeper to the first convolution layer and retrain the network with RECOLA dataset. Table~\ref{results_transfer_learning} shows that the highest CCCs are achieved by keeping frozen all the convolutional filters learned on the FER dataset and fine-tuning only the fully connected layers. Unfreezing convolutional layers reduces the CCC for both arousal and valence dimensions.

\begin{table}[htpb!]
\centering
\caption{CCC for the arousal and valence dimensions by fine-tuning none, one, two and all convolutional layers (CL) of the CNN.}
\renewcommand{\arraystretch}{1.2}
\label{results_transfer_learning}
\begin{tabular}{|c|c|c|c|c|}
\hline
\multirow{3}{*}{\textbf{Dimension}}  & \multicolumn{4}{c|}{\textbf{CCC}} \\ \cline{2-5}
& \multicolumn{4}{c|}{\textbf{Trainable CLs}} \\ \cline{2-5}
\multicolumn{1}{|c|}{} & \multicolumn{1}{c|}{\textbf{None}} & \multicolumn{1}{c|}{\textbf{3}} & \multicolumn{1}{c|}{\textbf{2, 3}} & \multicolumn{1}{c|}{\textbf{1, 2, 3}} \\ \hline
\multicolumn{1}{|l|}{Arousal} & \textbf{0.239} & 0.215 & 0.208 & 0.203 \\ \hline
\multicolumn{1}{|l|}{Valence} & \textbf{0.314} & 0.301 & 0.297 & 0.295 \\ \hline
\end{tabular}
\end{table}

\subsection{Pre-Processing}
The RECOLA dataset provides annotation for face localization within the video frames. However, for several video frames the annotation is not available. Even our face detector is not able to find the face in these video frames. Therefore, we can discard these frames since they may not contain a frontal face. However, if we drop all these frames, we end up discarding 22\% and 12\% of the video frames of the training and developing datasets, respectively. This allow us to improve the quality of the training data, nevertheless we reduce the amount of data for training the CNN. We introduced a second face detector 
that is able to retain 25\% of the data discarded previously. Besides that, we augment the remaining data using low-level transformations to reduce overfitting while training the CNN.


Table~\ref{results1} shows a summary of the best CCC values for arousal and valence achieved by fine-tuning the CNN. The results consider the impact of the delay compensation and the preprocessing (PP) on the CCC. We see that the preprocessing increases the CCC for both arousal and valence dimensions. Delays of 70 and 50 frames correspond to 2.8 and 2.0 seconds, respectively.

\begin{table}[htpb!]
\centering
\caption{The best CCC scores provided by the CNN for arousal and valence, for fine-tuning the convolutional layers (CLs), with delay compensation, and pre-processing (PP).}
\renewcommand{\arraystretch}{1.2}
 		\begin{tabular}{|c|c|c|c|c|c|c|c|}
 		\hline

{\bf Dimension} &{\bf Trainable CLs} &{\bf Delay} &{\bf PP} & {\bf CCC} \\

 	  \hline
			Arousal & 1, 2, 3 & 70 & Yes  & \textbf{0.252}\\
 	  \hline
 			Arousal & 1, 2, 3 & 70 & No & 0.239\\

 	  \hline
 			Arousal & None  & 70 & Yes & 0.203 \\
 	  \hline
      		Valence & 1, 2, 3 & 50 & Yes & \textbf{0.358} \\
 	  \hline
            Valence & 1, 2, 3 & 50& No &  0.314 \\
 	  \hline
       		Valence & None & 50 & Yes & 0.294\\
 	  \hline
 		\end{tabular}
        \label{results1}
\end{table}

\subsection{Multimodal Fusion}
In the proposed approach we use the CNNs only as feature extractors. Instead of selecting only the CNNs that lead to best CCCs (Table~\ref{results1}), we have selected also the CNNs that use transfer learning as initialization (no trainable CLs). Besides that, we have also considered the output of both the first (FC50) and the second (FC100) fully connected layers as feature representations. The idea behind that is to increase the diversity. The features extracted with the CNNs are then used to train an SVR.

\begin{table}[htpb!]
\centering
\caption{CCC for SVR trained on CNN features for arousal and valence dimensions.}
\label{tableA}
\renewcommand{\arraystretch}{1.2}
\begin{tabular}{|c|c|c|c|c|c|}
\hline

{\bf Dimension} &{\bf Trainable CLs} &{\bf Delay} & {\bf FC} & {\bf CCC} \hfill \\

 	  \hline
			Arousal & 1, 2, 3 & 70 &  100 & 0.146\\
 	  \hline
      	Arousal & 1, 2, 3 & 70 &  50 & {\bf 0.154} \\
 	  \hline
 			Arousal & None & 70 & 100 & 0.148 \\
 	  \hline
 			Arousal & None & 70 & 50 & 0.059 \\
 	  \hline
      		Valence & 1, 2, 3 & 50 & 100 & 0.429 \\
 	  \hline
   Valence & 1, 2, 3 & 50 & 50 & 0.414 \\
 	  \hline
       		Valence & None & 50& 100 & {\bf 0.433}\\
 	  \hline
       		Valence & None & 50 & 50 &0.414\\
 	  \hline
 		\end{tabular}
\end{table}

Table~\ref{tableA} shows the best CCCs achieved by the SVRs trained on CNN features. Reducing the dimension of the feature vector from 100 to 50 value leads to a reduction in CCC for valence but not for arousal. Furthermore, the best CCC achieved for arousal is for the features generated by fine-tuned CLs while for valence, the best CCC was achieved for features generated by the CNN with CLs learned on FER dataset. The reduced feature vectors preserve the essential information and generates better support vectors and therefore better predictions than a larger vector that may contain noisy features.

\begin{table}[htpb!]
\centering
\caption{Fusion results for \textbf{arousal} and \textbf{valence} with delay compensation of 70 and 50 respectively.}
\renewcommand{\arraystretch}{1.2}
\label{arousalfusionresults}
\begin{tabular}{|c|c|c|c|} 
\hline
\multicolumn{4}{|c|}{\textbf{Arousal}}\\
\hline
\textbf{Modality} &  \textbf{Feature} &  \textbf{Fusion}  & \textbf{CCC}  \\ 
\hline
Video & CNN [0CL, FC50]& \multirow{2}{*}{Early}& \multirow{2}{*}{\textbf{0.749}} \\ 
\cline{1-2}
Audio  &   Recola& & \\ 
\hline
Video & CNN [3CLs, FC50] &  \multirow{2}{*}{Early} & \multirow{2}{*}{\textbf{0.749}} \\ 
\cline{1-2}
Audio  &  Recola&  & \\ 
\hline
Video & CNN[3CLs, FC100]& \multirow{2}{*}{Late} & \multirow{2}{*}{0.701} \\ 
\cline{1-2}
Audio   & Recola &  & \\ 
\hline
Video &  CNN [3CLs, FC50] & \multirow{2}{*}{Late} & \multirow{2}{*}{0.715} \\ 
\cline{1-2}
Audio  &  Recola& & \\\hline
\hline
\multicolumn{4}{|c|}{\textbf{Valence}}\\
\hline
Video  & CNN [3CLs, FC100] & \multirow{2}{*}{Early}& \multirow{2}{*}{\textbf{0.565}} \\ 
\cline{1-2}
Audio  &Recola& & \\ 
\hline
Video  & CNN [0CL, FC100] & \multirow{2}{*}{Early} & \multirow{2}{*}{\textbf{0.551}} \\ 
\cline{1-2}
Audio  &Recola& & \\ 
\hline
Video  &CNN [0CL, FC100]&  \multirow{2}{*}{Late} & \multirow{2}{*}{0.543}  \\ 
\cline{1-2}
Recola& Audio  & & \\ 
\hline
Video  &CNN [3CLs FC50]& \multirow{2}{*}{Late} & \multirow{2}{*}{0.522}  \\ 
\cline{1-2}
Recola& Audio  & & \\
\hline
\multicolumn{4}{l}{\scriptsize CL: \# Frozen Convolutional Layers.}
\end{tabular}
\end{table}

\begin{figure*}[htpb!]
\centering
\includegraphics[width=0.75\textwidth]{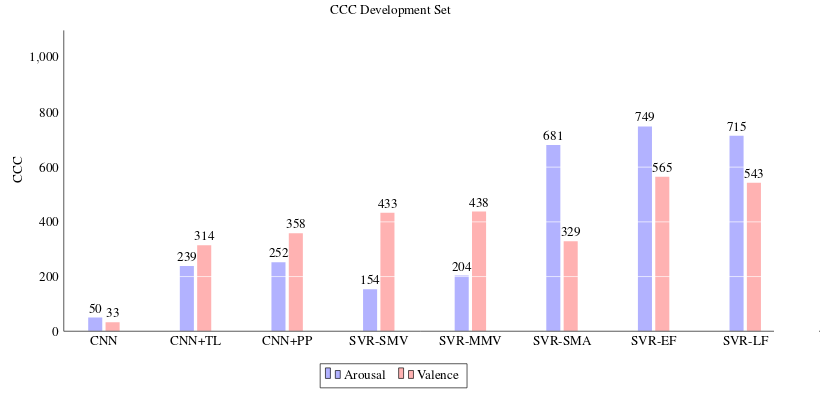}
\caption{Summary of the results by adding different elements: CNN alone; CNN with transfer learning and preprocessing (TL); CNN+TL plus post processing (CNN+PP); support vector regression with CNN features (SVR-SMV); support vector regression with audio features (SVR-SMA); multimodal early fusion (SVR-EF); multimodal late fusion (SVR-LF).}
\label{barras}
\end{figure*}

Table~\ref{arousalfusionresults} shows the CCCs for early and late multimodal fusion of CNN and audio features. It is important to highlight that SVRs trained on eGeMAPS audio features \cite{Valstar2016} achieved CCCs of 0.681 and 0.329 for arousal and valence, respectively. The results of Table~\ref{arousalfusionresults} show how complementary is the information between video and audio modalities and also, how each modality contributes to the final prediction. Fig.~\ref{barras} summarizes the changes of the CCC according to the inclusion of a new technique (transfer learning by using FER dataset, preprocessing, and post-processing). 

\section{Conclusion}
In this paper we have presented a multimodal approach for continuous emotions recognition that combines visual and acoustic information to predict arousal and valence levels of speakers. The proposed approach is based on a pre-trained CNN to extract features which are further combined with handcrafted audio features. The proposed approach outperforms most of the current approaches.  

The experimental results have shown the great impact of transfer learning in our model for the video modality. In the multimodal fusion, recent studies have shown that despite the fact that new and exotic fusion strategies are being developed, traditional fusion schemes are still able to  generate competitive results \cite{Cossetin2016,Duong2017,Li2017,Liang2018}.

\bibliographystyle{ieee}

\end{document}